\documentclass{article}

\PassOptionsToPackage{numbers, compress}{natbib}
 \usepackage[preprint]{neurips_2026}

\usepackage[utf8]{inputenc} 
\usepackage[T1]{fontenc}    
\usepackage[hidelinks]{hyperref}       
\usepackage{url}            
\usepackage{booktabs}       
\usepackage{amsfonts}       
\usepackage{amsmath}
\usepackage{amssymb}
\usepackage{nicefrac}       
\usepackage{microtype}      
\usepackage{xcolor}         
\usepackage[pdftex]{graphicx}
\usepackage{subcaption}
\usepackage{mathtools}
\usepackage{multirow}
\usepackage{tabularx}
\usepackage{wrapfig}

\usepackage{todonotes}
\makeatletter
\newcommand*\iftodonotes{\if@todonotes@disabled\expandafter\@secondoftwo\else\expandafter\@firstoftwo\fi} 
\makeatother

\title{Tracing Uncertainty in Language Model ``Reasoning''}

\author{%
  \textbf{Nils Grünefeld\textsuperscript{1,4,$\star$}} \quad
  \textbf{Bertram Højer\textsuperscript{1,4,$\star$}} \quad
  \textbf{Philipp Mondorf\textsuperscript{2,5}} \\
  \textbf{Barbara Plank\textsuperscript{2,5}}\,
  \textbf{Anna Rogers\textsuperscript{1,4}}\,
  \textbf{Christian Hardmeier\textsuperscript{1,4}}\,
  \textbf{Stefan Heinrich\textsuperscript{1,4}}\,
  \textbf{Jes Frellsen\textsuperscript{3,4}} \\[0.4em]
  \textsuperscript{1}Data Science Section, IT University of Copenhagen \\
  \textsuperscript{2}MaiNLP, Center for Information and Language Processing, LMU Munich \\
  \textsuperscript{3}Department of Applied Mathematics and Computer Science, Technical University of Denmark \\
  \textsuperscript{4}Pioneer Centre for Artificial Intelligence \quad
  \textsuperscript{5}Munich Center for Machine Learning (MCML) \\[0.3em]
  \textsuperscript{$\star$}Equal contribution; order settled by coin-flip.
}

\begin{document}

\newcommand{\UTA}{\mathbb{U}^{\mathrm{A}}}
\newcommand{\UT}{\mathbb{U}^{\mathrm{Tr}}}
\newcommand{\UE}{\mathbb{U}_{\mathrm{E}}}
\newcommand{\UA}{\mathbb{U}_{\mathrm{C}}}
\newcommand{\UH}{\mathbb{U}_{\mathrm{D}}}

\maketitle

\begin{abstract}
Language model (LM) ``reasoning'', commonly described as Chain-of-Thought or test-time scaling, often improves benchmark performance, but the dynamics underlying this process remain poorly understood.
We study these dynamics through the lens of uncertainty quantification by treating the ``reasoning'' traces, the intermediate token sequences generated by LMs, as evolving model states.
We summarize each trace by an \emph{uncertainty trace profile}: a small set of features describing the shape of the uncertainty signal over its trace, such as its slope and linearity.
We find that across five LMs evaluated on GSM8K and ProntoQA, these profiles predict whether a trace yields a correct final answer with AUROC up to $0.807$, improving markedly on recent related work. We reach AUROC $0.801$ using only the first few hundred tokens of full traces, suggesting that errors can be detected \textit{early} in the generation.
A detailed comparison of correct and incorrect traces further reveals qualitatively distinct uncertainty profiles, with correct traces showing a steeper and less linear decline in uncertainty.
Together, the results suggest that our method, grounded in decision-making under uncertainty, provides a principled lens for studying the generative process underlying LM ``reasoning''.
\end{abstract}

\section{Introduction}
\label{sec:intro}


The notion of \textit{reasoning} is increasingly being used to describe the output generated by language models (LMs).
Whether referred to as a ``scratchpad'' \cite{nye-2021-showyour}, ``Chain-of-Thought'' (CoT) \cite{wei-2022-chainofthoughtprompting, kojima-2022-largelanguage}, or test-time scaling \cite{muennighoff-2025-s1simple}, researchers generally refer to the same process: an LM generating a long sequence of intermediate tokens that---ideally---results in a correct final answer.
It is currently being debated whether ``reasoning'' is the right term to describe this process \cite{stechly-2024-chainthoughtlessness, stechly-2025-semanticsunreasonable, kambhampati-2025-stopanthropomorphizing, hojer-2025-notionthat}, but irrespective of that debate, open-ended long-form generation has been shown to improve performance on a variety of benchmarks with verifiable labels such as mathematical and logical reasoning \cite{yang-2025-qwen3technical, guo-2025-deepseekr1incentivizes}.
The connection between the length of the CoT and task performance is mainly correlational \cite{liu-2025-understandingr1zerolike}, although some work suggests that intervening on the ``thinking'' budget can directly affect performance \cite{muennighoff-2025-s1simple}.

``Reasoning under uncertainty'' is a well-established concept in the psychology of decision-making, which treats reasoning and decision-making as the process of forming judgments and selecting among alternatives, often given incomplete knowledge of outcomes and specific probabilities \cite{tversky-1974-judgmentuncertainty, tversky-1981-framingdecisions, tversky-1986-rationalchoice}.
A definition of reasoning still eludes NLP and AI, and labeling the generative process of LMs as such arguably reflects misguided anthropomorphization.
Nevertheless, the notion of reasoning under uncertainty is particularly relevant for this paper.
Broadly speaking, definitions of \textit{reasoning} in NLP and AI describe \textit{some} process that is ``\textit{logical and systematic}'' \cite{wei-2022-chainofthoughtprompting, kojima-2022-largelanguage, huang-2023-reasoninglarge}, but even this vague ``definition'' is questionable.
Generally, LMs do not enforce the logical validity of generated outputs, and the reinforcement learning (RL) optimization explicitly rewards only the generation of a correct answer, not the validity of the reasoning trace \cite{guo-2025-deepseekr1incentivizes, yang-2025-qwen3technical}.

\begin{figure}[t!]
    \centering
    \includegraphics[width=1\linewidth]{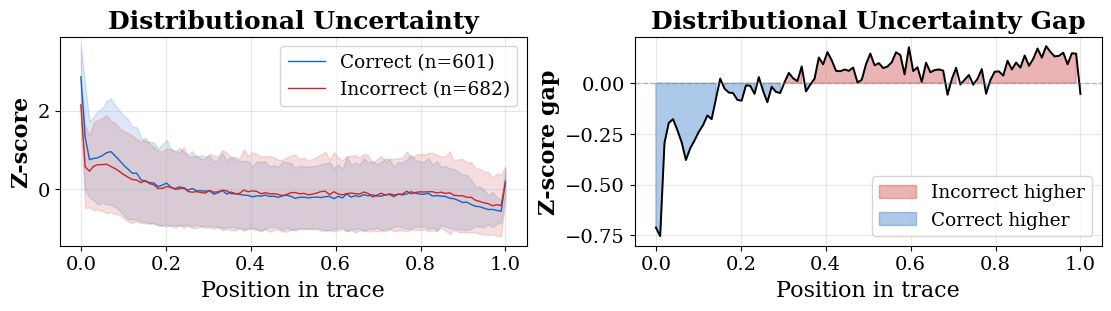}
    \caption{The average uncertainty trace for Qwen 2.5 on GSM8K. Left illustrates the average distributional uncertainty (entropy) computed over all generated sequences. Right shows the average uncertainty gap between correct and incorrect sequences.}
    \label{fig:avg-uncertainty-traces}
\end{figure}

In this paper, we treat the generated CoT of an LM as an evolving state of the model;
a state which can be more or less certain.
\autoref{fig:avg-uncertainty-traces} illustrates this perspective: on the left we see the normalized average of the \textit{distributional uncertainty} (entropy) computed at every step of generation.
While correct and incorrect traces overlap and show a similar decrease in uncertainty as a trace progresses, the gap plot (right) indicates sections of the averaged traces where correct and incorrect traces clearly deviate;
it is this deviation in the uncertainty dynamics we utilize.
We apply uncertainty quantification (UQ) to compute three measures of uncertainty to study how the model state develops as a time series, and show that the resulting \textbf{uncertainty trace profile} is strongly predictive of final answer correctness for mathematical and logical reasoning tasks.

\paragraph{Related work.}
\label{sec:intro:related}
A growing body of work has explored methods for improving LM performance at inference time via various interventions and uses of test-time compute, often by sampling multiple outputs and selecting among them.
\citet{wang-2023-selfconsistencyimproves} propose \textit{Self-Consistency} which samples CoT reasoning paths, selecting the most frequent answer by majority voting.
\textit{Universal self-consistency} \cite{chen-2023-universalselfconsistency} extends this approach to open-ended tasks by prompting the model itself to identify the most consistent response.
To address limitations related to scale, \citet{kang-2025-scalablebestofn} propose \textit{Self-Certainty}, a metric based on the Kullback-Leibler divergence of each token's predicted distribution from a uniform distribution, enabling best-of-N selection yielding better scaling as well as generalization to open-ended generation.
Similarly, \citet{fu-2025-deepthink} introduce \textit{DeepConf}, a sampling method that uses model-internal confidence signals to filter low-quality reasoning traces during or after generation.
Other approaches rely on hidden state representations to improve performance on verifiable tasks \cite{hojer-2025-improvingreasoning, zhang-2025-reasoningmodels} and \citet{zhao-2025-verifyingchainofthought} propose \textit{Circuit-Based Reasoning Verification} (\textit{CRV}) using transducers and the computation graph of Llama-3.1-8B to distinguish correct from incorrect generations.

\paragraph{Novelty and contribution.}
The aggregation inherent to other methods (\textit{CRV}, \textit{Self-Certainty}) discards most of the structure of the generative process; we treat per-token uncertainty as a time series, enabling richer characterization of how and where reasoning quality changes over the course of generation.
Our work is thus complementary to said approaches: like \textit{Self-Certainty} and \textit{DeepConf}, we leverage uncertainty signals from model outputs and internals, but we are the first to utilize the trace as an evolving state as illustrated in \autoref{fig:avg-uncertainty-traces}.
Our primary research questions are thus: \textbf{RQ1)} Are uncertainty trace profiles predictive of correctness; and \textbf{RQ2)} What are the signals of uncertainty that distinguish incorrect from correct generations?

\section{Background}
\label{sec:bg}

\subsection{``Reasoning'' in Language Models}
\label{sec:bg:reasoning-in-lms}


To formalize the mechanics of LM generation, we consider an LM as a parameterized function \mbox{$f_\theta : \mathcal{V}^{\leq K} \to \Delta(\mathcal{V})$}, where $\mathcal{V}$ is the vocabulary, $\mathcal{V}^{\leq K}$ denotes a sequence of up to $K$ tokens, and $\Delta(\mathcal{V})$ denotes the probability simplex over the vocabulary.
The model parameters $\theta$ define a conditional distribution $p(v_t \mid v_{<t}, \theta)$ over the next token $v_t \in \mathcal{V}$ given the prefix $v_{<t} = (v_1, \ldots, v_{t-1})$, from which outputs are sampled during generation.

Generating from this model autoregressively is the foundation for producing long sequences of intermediate tokens that simulate reasoning.
The finding that such intermediate generations improve performance on some verifiable tasks has resulted in the development of ``reasoning models'' (RMs) \cite{yang-2025-qwen3technical, guo-2025-deepseekr1incentivizes}.
In this paradigm a model is trained with RL to generate an extensive reasoning trace before generating the final answer.
Recent discussions emphasize that RMs are post-trained with both a different style of data as well as learning objective, such as RL with verifiable feedback (RLVF).
Some argue that models with this additional complexity are therefore necessarily not \textit{just} ``next-token predictors'' \cite{downes-2024-llmsare}, while others question how different RLVF is from standard optimization techniques \cite{shanahan-2023-talkinglarge, samineni-2025-rlname}.
Irrespective of this debate, the resulting models generate thousands of tokens and achieve improved performance on some verifiable tasks.

At surface-level, both LMs and RMs generate sequences of human-like reasoning.
But while longer generated traces are correlated with improved performance on some benchmarks, the logical validity, reliability, and faithfulness of ``reasoning'' traces have been contested \citep{lanham-2023-measuringfaithfulness, turpin-2023-languagemodels, mirzadeh-2025-gsmsymbolicunderstanding, mondorf-2024-comparinginferential, arcuschin-2025-chainofthoughtreasoning, chen-2025-reasoningmodels}.
This calls into question approaches that analyze RM outputs through their semantic content alone \cite{ling-2023-deductiveverification}, especially given evidence that training on semantically invalid CoT traces can still improve performance \cite{stechly-2025-semanticsunreasonable}.

\subsection{Uncertainty in Machine Learning}
\label{sec:bg:uncertainty-in-ml}

Most machine learning is inherently probabilistic;
even models that produce deterministic point predictions are implicitly embedded in a probabilistic framework \citep{mackay-1992-practicalbayesian, neal-2012-bayesianlearning, blundell-2015-weightuncertainty}.
Within this framework, two kinds of uncertainty are commonly differentiated: aleatoric and epistemic \cite{HORA1996217,KIUREGHIAN2009105,hullermeier-2021-aleatoricepistemic}.
\begin{enumerate}
    \item \textit{Aleatoric} uncertainty reflects the inherent, irreducible difficulty of the task itself: variability that no amount of further training could remove.
    \item \textit{Epistemic} uncertainty reflects the degree to which a model is informed by its training data about a particular prediction;
a decrease over the course of generation can thus be interpreted as ``being closer'' to regions of the output space encountered during training.
\end{enumerate}

\paragraph{Uncertainty quantification (UQ).}
Quantifying these uncertainties is commonly approached using information-theoretic measures, specifically through an entropy decomposition of the predictive distribution $H(y \mid x) = H(y \mid x, \theta) + I(y;\, \theta)$, where $x$ denotes the input, $y$ the prediction, $\theta$ the model parameters, and the two components on the right represent aleatoric and epistemic uncertainty, respectively \cite{smith-2018-understandingmeasures}.

Recently, this decomposition has been subject to criticism \citep{wimmer-2023-quantifyingaleatoric, tomov-2025-entropynot} and a label-wise variance-based alternative has been proposed \citep{sale-2023-secondorderuncertainty, sale-2024-labelwisealeatoric}.
This alternative leverages the law of total variance to decompose the total uncertainty for a given label $k$, defined as $\operatorname{Var}[Y_k]$, into an epistemic component $\operatorname{Var}_\theta[p(y_k \mid x, \theta)]$, capturing the variability of the predictive probability across different parameterizations, and an aleatoric component $\mathbb{E}_\theta[p(y_k \mid x, \theta)(1 - p(y_k \mid x, \theta))]$, capturing the expected inherent randomness of the prediction under a given parameterization.
For efficient estimation of epistemic uncertainty in LMs, we adopt a recently proposed isotropic approximation \citep{grunefeld-2026-isotropicapproach}, which to our knowledge is the only computationally tractable approach for variance-based uncertainty quantification at this scale, particularly for the epistemic component.

In this work, we primarily use the variance-based framework for the epistemic component, since its information-theoretic analogue---the mutual information $I(y;\,\theta)$---has no tractable estimator for models on the scale of LMs.
On the aleatoric side, we use both frameworks to further distinguish between two sub-types, committal and distributional:
\begin{enumerate}
    \item[1a.] \textit{Committal} aleatoric uncertainty, $\mathbb{E}_\theta[p(\hat{y}_k \mid x, \theta)(1 - p(\hat{y}_k \mid x, \theta))]$, the expected Bernoulli variance over $\theta$ of the predicted class $\hat{y}_k$, taken from the variance-based decomposition.
    \item[1b.] \textit{Distributional} aleatoric uncertainty, $\mathbb{E}_\theta[H(y \mid x, \theta)]$, the expected conditional entropy of the predictive distribution over $\theta$, taken from the information-theoretic decomposition.
\end{enumerate}
Committal uncertainty depends only on the probability $p$ assigned to the selected token, and distinguishes two qualitatively different ways in which a token can be generated.
When $p$ is near $1$, the model is committed to the specific token, in what we call a process of \textbf{concentration};
when $p$ is near $0.5$, committal uncertainty is at its maximum and the top token is selected not by being likely, but by being marginally less unlikely than its alternatives, in what we call a process of \textbf{elimination}.
Distributional uncertainty, by contrast, is sensitive to the shape of the entire predictive distribution rather than just the probability of its mode.
The two measures can therefore differ: a model may be firmly in the concentration regime, yielding low committal uncertainty, while still exhibiting high distributional uncertainty if the remaining mass is spread diffusely across many alternatives.
We provide intuition for these uncertainty types and illustrative examples in \autoref{app:uncertainty-intuitions}.
\section{Methods}
\label{sec:methods}

We analyze the development of uncertainty over the course of a trace as a novel approach to studying the generative process, in contrast to prior work which discards most of the temporal structure of autoregressive generation.

\subsection{Uncertainty Estimation}
\label{sec:methods:uncertainty}

Following the framework of \autoref{sec:bg:uncertainty-in-ml}, we estimate the three uncertainty types at every token position.
As described in \autoref{sec:bg:reasoning-in-lms}, the autoregressive generation is conditioned on the trace prefix $v_\mathit{<t}$ and we let $y$ be the target.
The choice of target distinguishes two settings: for \textbf{trace uncertainty} the target is the next token $v_\mathit{t}$, while for \textbf{answer uncertainty} it is the eventual final answer generated by the model, $\hat{y}$.

The epistemic component is estimated using the squared L2 norm of the gradient of the predicted probability of the target with respect to all model parameters, derived via a first-order delta-method expansion of $\operatorname{Var}_\mathit{\theta}[p(y \mid v_\mathit{<t}, \theta)]$ around a point estimate $\hat{\theta}$ of the parameters \citep{grunefeld-2026-isotropicapproach}:
\begin{equation}
    \UE \coloneq \operatorname{Var}_\theta[p(y \mid v_{<t}, \theta)] \approx \left\| \nabla_\theta \, p(y \mid v_{<t}, \theta) \, \big|_{\theta=\hat{\theta}} \right\|^2.
\end{equation}

\textit{Committal aleatoric uncertainty} is estimated as the Bernoulli variance of the target's probability,
\begin{equation}
    \UA
    \coloneq \mathbb{E}_\theta[p(y \mid v_{<t}, \theta)(1 - p(y \mid v_{<t}, \theta))]
    \approx p(y \mid v_{<t}, \hat{\theta}) \cdot (1 - p(y \mid v_{<t}, \hat{\theta})).
\end{equation}

\textit{Distributional aleatoric uncertainty} is computed as the predictive entropy over the full vocabulary, conditioned on the prefix and evaluated at the point estimate $\hat{\theta}$:
\begin{equation}
    \UH \coloneq \mathbb{E}_\theta[H(\,\cdot \mid v_{<t}, \theta)] \approx H(\,\cdot \mid v_{<t}, \hat{\theta}) = - \sum_{v \in \mathcal{V}} p(v \mid v_{<t}, \hat{\theta}) \log p(v \mid v_{<t}, \hat{\theta}).
\end{equation}

\paragraph{Trace and answer uncertainty.}
At each step $t$, trace uncertainty, $\UT$, instantiates the three measures above with $y = v_t$, characterizing the state of a model with respect to the token it generates next.
Answer uncertainty, $\UTA$, instantiates them with $y = \hat{y}$, characterizing instead the state of a model with respect to the eventual answer; this requires a second forward and backward pass at each step, conditioned on the trace prefix $v_\mathit{<t}$ and applied to $\hat{y}$ rather than $v_\mathit{t}$ (see \autoref{app:splitting}) \cite{grunefeld-2026-isotropicapproach}.
Since $\hat{y}$ is in general a multi-token span, we use the sequence-level extensions of \citet{grunefeld-2026-isotropicapproach} for epistemic and committal uncertainties, and analogously the average entropy for distributional uncertainty.

We refer to $\UE$, $\UA$, and $\UH$ as the three \textbf{types} of uncertainty, and to $\UT$ and $\UTA$ as the two uncertainty \textbf{channels}.
The two are orthogonal: each type has a trace variant and an answer variant, giving six time series per generation that together describe how local and answer-directed uncertainty co-evolve.

\subsection{Experimental Procedure}
\label{sec:methods:procedure}

\paragraph{Models \& datasets.}
\label{sec:methods:models}

We evaluate Llama-3.1 (\textit{8B}), Llama-3.2 (\textit{1B}), Qwen-2.5 (\textit{0.5B}), DeepSeek R1 Distill Qwen (\textit{1.5B}), and Qwen-3 (\textit{0.6B}).
These models are chosen for comparability with related work \cite{zhao-2025-verifyingchainofthought} and to represent different training techniques.
Llama 3 models are trained primarily using supervised fine-tuning and limited RL with human feedback and thus without explicit ``reasoning'' optimization \cite{grattafiori-2024-llama3}.
Qwen-2.5 was trained similarly, but with a focus on ``reasoning'' data \cite{yang-2024-qwen2technical}, whereas DeepSeek R1, and Qwen-3 were trained with the explicit objective of generating ``reasoning'' traces to improve correctness \cite{guo-2025-deepseekr1incentivizes, yang-2025-qwen3technical}.
These models range from standard LMs (the non-``reasoning'' type) (Llama) to RMs (the ``reasoning''-type) (R1, Qwen3), with Qwen-2.5 landing in-between.
This enables an assessment of how uncertainty dynamics differ on this spectrum.

We evaluate on two datasets: GSM8K \cite{cobbe-2021-trainingverifiers}, chosen for comparability with related work \cite{zhao-2025-verifyingchainofthought}, and ProntoQA \cite{saparov-2022-languagemodels} as a separate benchmark.
GSM8K consists of grade school word math problems of varying difficulty while ProntoQA consists of small logical problems using low-likelihood tokens.
GSM8K and ProntoQA are examples of benchmarks with verifiable answers on which RMs have shown strong performance improvements compared to LMs.
We further study examples from GSM-Symbolic \cite{mirzadeh-2025-gsmsymbolicunderstanding}; a variant of GSM8K with controlled perturbations that preserve the underlying logic of questions.

\paragraph{Generating traces.}
We evaluate each model on both datasets using greedy decoding.
The generated outputs are split into a ``reasoning'' trace and a final answer using model-specific splitting strategies that are documented in \autoref{app:splitting}.
We initially set a max sequence length of $2048$ for the trace and filter out responses that do not produce a final answer.
For the reasoning models we append the \texttt{</think>} token to any trace that did not generate a final answer within the initial generation limit, restart the generation with an additional $512$ token budget, and again filter out the traces that did not generate a final answer (details in \autoref{app:experimental-setup});
the filtering removed approximately $10\%$ of samples for DeepSeek R1 and Qwen 3.

\subsection{Predictive Modeling \& Reasoning Dynamics}
\label{sec:methods:modeling}

\paragraph{Feature engineering \& correctness prediction.}
We extract a small set of features from all type-channel combinations ($\UH^{\mathrm{Tr}}, \UA^{\mathrm{Tr}}, \UE^{\mathrm{Tr}}, \UH^{\mathrm{A}}, \UA^{\mathrm{A}}, \UE^{\mathrm{A}}$) to summarize the dynamics of a trace into the uncertainty trace profile.
We compute the early mean ($\mu_{\mathrm{early}}$), the middle mean ($\mu_{\mathrm{mid}}$), the late mean ($\mu_{\mathrm{late}}$), the linear slope of the trace ($\mathbf{m}$), and the fit of the slope ($r^2$), resulting in $30$ features in total (see \autoref{app:trace-feats}; \autoref{tab:trace-features}).
Using these features, we train two binary classifiers to predict the correctness of the final answer of every given trace, for each model-dataset combination. We choose a logistic regression (LogReg) classifier for its interpretability, and a gradient boosted model (GBoost) for comparability with related work \cite{zhao-2025-verifyingchainofthought}. Full experimental details are provided in \autoref{app:experimental-setup}.
We use the area under the receiver operating characteristic curve (AUROC) to measure the predictive performance of each classifier in a five-fold cross-validation setup.

\paragraph{Early correctness detection.}
\label{sec:methods:early-detection}
To assess how early we can predict correctness, we bin traces by number of tokens generated, and extract features based only on the tokens of the trace before the bin limit.
We train the same three classifiers based on the constrained traces, to analyze how informative early trace dynamics are for predicting correctness.
\section{Analysis}
\label{sec:analysis}

Using the uncertainty trace profile to predict correctness improves on the AUROC scores of \textit{CRV} \cite{zhao-2025-verifyingchainofthought} by $11.5\%$ and \textit{Self-Certainty} \cite{kang-2025-scalablebestofn} by up to $59\%$.
We report AUROC scores across models and datasets, and perform an in-depth an analysis of the uncertainty trace profile for correct and incorrect traces (i.e. traces that result or don't result in a correct answer).

\subsection{Correctness Prediction}
\label{sec:analysis:correctness}

\autoref{tab:auroc-combined} displays the correctness prediction AUROC scores for each model-dataset combination.
We report the scores achieved with our uncertainty trace profiles using both LogReg and GBoost classifiers as well as the \textit{Self-Certainty} approach for comparison.
Across all models, our uncertainty-based approach matches or outperforms \textit{Self-Certainty}, and most of the predictive power is carried by $\UT$ (see \autoref{app:analysis}; \autoref{tab:auroc-split}).
Notably, the uncertainty trace profile is predictive of correctness regardless of model type, in contrast to \textit{Self-Certainty}.
Further, we improve on the state-of-the-art \textit{CRV} approach \cite{zhao-2025-verifyingchainofthought}, which evaluated a single model, Llama-3.1-8B on GSM8K, and reached an AUROC of $0.702$ versus our $0.783$.

\begin{table}[h!]
\centering
\small
\begin{tabularx}{\textwidth}{l *{6}{>{\centering\arraybackslash}X}}
\toprule
& \multicolumn{3}{c}{GSM 8K} & \multicolumn{3}{c}{ProntoQA} \\
\cmidrule(lr){2-4} \cmidrule(lr){5-7}
& LR & GB & \textit{SC} & LR & GB & \textit{SC} \\
\midrule
Llama 3.1   & \textbf{0.783} {\tiny ±0.031} & 0.758 {\tiny ±0.026} & 0.491 {\tiny ±0.063} & \textbf{0.799} {\tiny ±0.052} & 0.762 {\tiny ±0.101} & 0.566 {\tiny ±0.080} \\
Llama 3.2   & 0.758 {\tiny ±0.017} & \textbf{0.767} {\tiny ±0.036} & 0.566 {\tiny ±0.041} & \textbf{0.550} {\tiny ±0.061} & 0.533 {\tiny ±0.073} & 0.549 {\tiny ±0.065} \\
Qwen 2.5    & \textbf{0.807} {\tiny ±0.017} & 0.787 {\tiny ±0.016} & 0.689 {\tiny ±0.028} & 0.519 {\tiny ±0.034} & 0.476 {\tiny ±0.050} & \textbf{0.565} {\tiny ±0.052} \\
DeepSeek R1 & \textbf{0.786} {\tiny ±0.045} & 0.775 {\tiny ±0.044} & 0.703 {\tiny ±0.028} & \textbf{0.672} {\tiny ±0.081} & 0.639 {\tiny ±0.106} & 0.615 {\tiny ±0.083} \\
Qwen 3      & 0.727 {\tiny ±0.057} & 0.665 {\tiny ±0.040} & \textbf{0.728} {\tiny ±0.036} & \textbf{0.657} {\tiny ±0.078} & 0.551 {\tiny ±0.051} & 0.611 {\tiny ±0.153} \\
\bottomrule
\end{tabularx}
\caption{\small AUROC scores (5-fold CV) across five models and classifiers for GSM8K and ProntoQA. Values shown as mean ± std.
LR = Logistic Regression, GB = Gradient Boosting, \textit{SC} = \textit{Self-Certainty}.}
\label{tab:auroc-combined}
\end{table}

\paragraph{Uncertainty type classifiers.}
We train classifiers for the trace and answer channels separately and observe similar, although slightly lower AUROC scores on the answer channel (see \autoref{app:analysis}; \autoref{tab:auroc-split}).
We further train individual LogReg models for each type-channel combination, observing similar scores for $\UH$ and $\UA$ (AUROC up to $0.78$) and slightly lower scores for $\UE$ (\autoref{app:analysis}; \autoref{tab:auroc-uncertainty-types}).
This is likely due to the fact that $\UA$ and $\UH$ are similar: they both approximate aspects of aleatoric uncertainty and are thus closer to each other than to their epistemic counterpart.

\begin{figure}[h!]
    \centering
    \includegraphics[width=.95\linewidth]{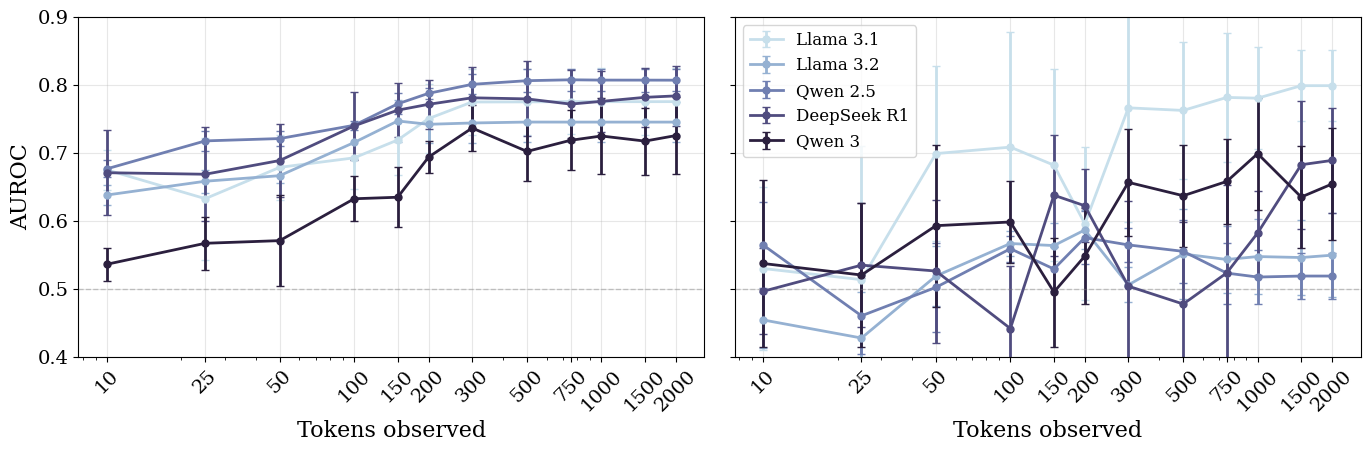}
    \caption{\small AUROC scores on models trained on features extracted based on increasing shares of the full traces (left: GSM8K, right: ProntoQA).}
    \label{fig:trace-streaming-absolute}
\end{figure}

\paragraph{Early correctness detection.}
\autoref{fig:trace-streaming-absolute} shows that the UQ lens and our uncertainty trace profiles are informative of correctness relatively early in generation for trace-level uncertainty, $\UT$.
We observe AUROC up to $0.801$ at the $300$th token for GSM8K, and a similar trend across all five models, with AUROC rising steadily as larger parts of the full trace are included.
A similar trend is visible for ProntoQA, although results are more noisy.
While we primarily focus on the dynamics of uncertainty traces over the course of generation, this result indicates that our approach could be used to resample traces even before generation halts, improving on the efficacy of resampling approaches.

\paragraph{Feature importance.}

\begin{wrapfigure}{r}{0.5\textwidth}
    \centering
    \includegraphics[width=0.5\textwidth]{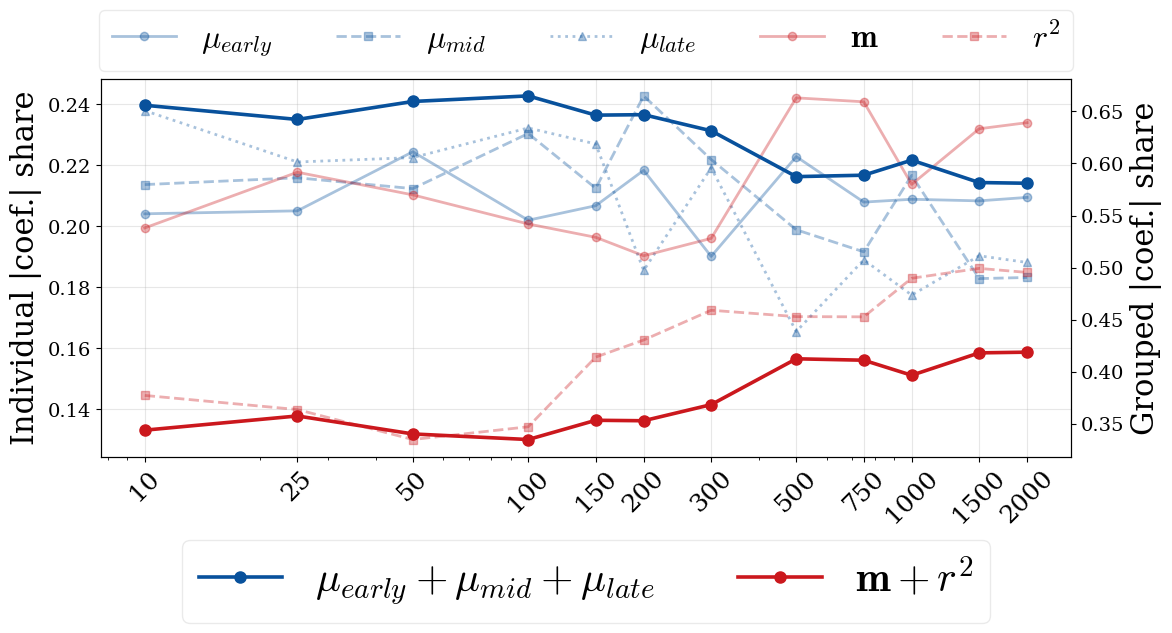}
    \caption{\small Change in coefficient share over the course of generation.}
    \label{fig:feature-importance}
\end{wrapfigure}

We measure feature importance via the coefficient share of the logistic regression, which is computed as the absolute coefficient of each feature as a fraction of the sum of all absolute coefficients.
We aggregate static ($\mu_{\mathrm{early}}, \mu_{\mathrm{mid}}, \mu_{\mathrm{late}}$) and dynamic (slope, $r^2$) features to assess the group-level importance, which is computed by summing the coefficient shares in each group. \autoref{fig:feature-importance} shows how the coefficient shares change over the course of a trace:
at the very beginning of a generated trace the static uncertainty markers have the largest coefficient share, but as the generated sequences become longer, the slope and linearity coefficients of the uncertainty trace profile become more important in predicting correctness.

\subsection{Feature Analysis}
\label{sec:analysis:features}

\begin{figure}[b]
    \centering
    \includegraphics[width=1\linewidth]{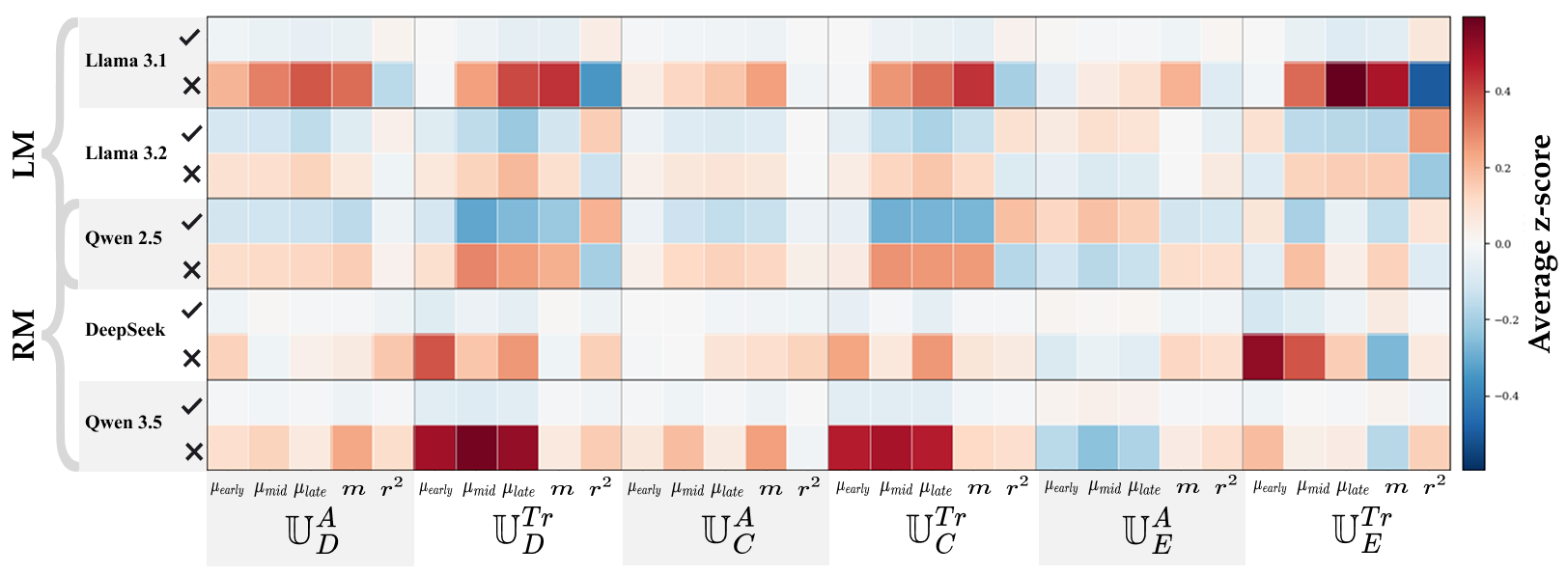}
    \caption{Heatmap of uncertainty features across models and datasets. The heatmap is organized into six feature blocks---$\UH$, $\UA$, and $\UE$ for $\UTA$ and $\UT$ respectively---each containing five features: $\mu_{\mathrm{early}}$, $\mu_{\mathrm{mid}}$, $\mu_{\mathrm{late}}$, slope ($\mathbf{m}$), and $r^2$. Colors indicate relative differences within model between correct and incorrect traces.}
    \label{fig:feature-heatmap}
\end{figure}

We examine the per-model uncertainty trace profiles shown in \autoref{fig:feature-heatmap}, which displays features for incorrect and correct traces standardized by their z-score to focus on relative differences between correct and incorrect traces.
We standardize within model because the absolute magnitudes of two of our measures are not informative for cross-model comparison.
Gradient-based epistemic uncertainty, $\UE$, is not directly comparable across models and entropy, $\UH$, is comparable in principle but dominated in scale by the long tail of effectively irrelevant tokens.

\paragraph{Levels.}
Across all five models, $\mu_{\mathrm{early}}$, $\mu_{\mathrm{mid}}$, and $\mu_{\mathrm{late}}$ of most uncertainty types are higher for incorrect traces than for correct ones: the static level of uncertainty points to an incorrect answer as can be seen by comparing the rows of \autoref{fig:feature-heatmap}.
The gap is more pronounced for trace-level uncertainty, $\UT$, than for answer-level uncertainty, $\UTA$, consistent with the finding that $\UT$ features carry most of the predictive power (\autoref{app:analysis}; \autoref{tab:auroc-split}).
The clear exception is answer-level epistemic uncertainty, $\UE^{\mathrm{A}}$, where we observe the opposite, with incorrect traces showing lower uncertainty.
The two epistemic measures therefore point in opposite directions on failure.
$\UE^{\mathrm{Tr}}$ is elevated on incorrect traces, indicating that the trace-level steps of an incorrect trace are less informed by training data than those of a correct trace;
$\UE^{\mathrm{A}}$ is depressed on incorrect traces, indicating that the eventual wrong answer is more informed by training data than correct answers tend to be.
Our results can be interpreted as follows: when a model generates an incorrect answer, it generates less-supported steps to arrive at a wrong answer that is nonetheless closer to regions encountered during training.
This pattern is complementary to the elevated aleatoric uncertainty measures observed on the same incorrect traces.

\paragraph{Temporal dynamics.}
Slope ($\mathbf{m}$) and $r^2$ characterize how the uncertainty trajectory develops over the course of generation as illustrated by the columns of \autoref{fig:feature-heatmap}.
$\mathbf{m}$ distinguishes incorrect from correct traces consistently across models: in almost all cases, the slope of incorrect traces is higher relative to correct traces, indicating that uncertainty decreases less in incorrect traces as the generation progresses.
The exception is $\UE^{\mathrm{Tr}}$: in Llama 3.1 the slope is still higher for incorrect traces, but the gap shrinks as we move from LMs and inverts in RMs, so that for DeepSeek-R1 and Qwen 3 the $\UE^{\mathrm{Tr}}$ slope of incorrect traces is again lower.

The fit quality, $r^2$, follows a similar trend.
In Llama 3.1, almost every uncertainty type has higher $r^2$ for correct traces than for incorrect ones, with correct trajectories following tighter linear trends and incorrect trajectories appearing more irregular.
Moving from LMs to RMs this pattern flips such that incorrect traces appear more linear than correct ones.

In Llama 3.1, the local steps of incorrect traces are persistently less informed by training data than those of correct traces, with $\UE^{\mathrm{Tr}}$ elevated throughout and not decreasing as the trace progresses.
The eventual wrong answer is nonetheless more strongly supported by training data than correct answers tend to be.
The trace and the answer it produces thus pull in opposite directions.
In the RMs the picture changes: $\UE^{\mathrm{Tr}}$ decreases more steeply on incorrect traces than on correct ones, so the local steps move progressively closer to better-supported territory as the model converges on the wrong answer that $\UE^{\mathrm{A}}$ already flagged as well-supported.
The smoother, more linear $r^2$ profiles of incorrect RM traces are consistent with this convergence.
Together, the level and temporal dynamics describe a qualitative shift in how incorrect traces fail across model classes.

\subsection{Qualitative Analysis}
\label{sec:analysis:qualitative}

\begin{figure}[b]
    \centering
    \includegraphics[width=1\linewidth]{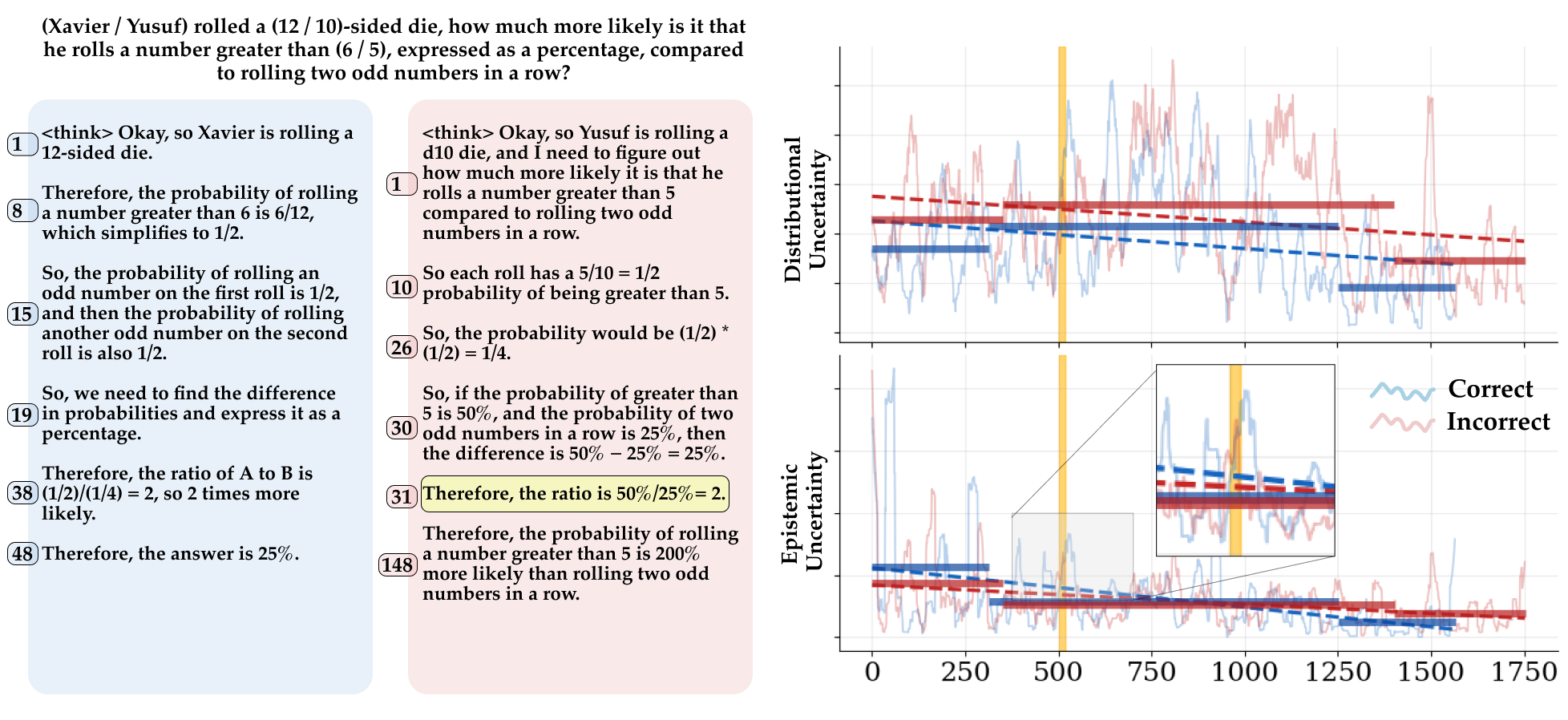}
    \caption{\small Sample from GSM-Symbolic. Numbers indicate sentence indices. $\mu_{\mathrm{early}}$, $\mu_{\mathrm{mid}}$, and $\mu_{\mathrm{late}}$ are shown as horizontal lines and slope as a dashed linear-fit line, with the corresponding $r^2$ reflected in the spread of the data around that line. Each plot shows the uncertainty type with the token index on the x-axis, and the location of the first error in the incorrect trace is highlighted in yellow.}
    \label{fig:gsm-symbolic}
\end{figure}

To illustrate qualitatively how the analysis of uncertainty can be applied to study LM ``reasoning'', we select a pair of sequences generated by Qwen 3 when evaluated on a GSM-Symbolic example \citep{mirzadeh-2025-gsmsymbolicunderstanding}.
Recall that GSM-Symbolic examples are perturbed GSM8K questions that preserve the underlying logic.
\autoref{fig:gsm-symbolic} shows an example where the name and the size of the die differ between questions.
We show the generated sequences (left) alongside the smoothed trace-level trajectories of distributional uncertainty $\UH^{\mathrm{Tr}}$ and epistemic uncertainty $\UE^{\mathrm{Tr}}$, with the uncertainty trace profile features overlaid on the raw trace values (right): horizontal lines show the three means, dashed lines show the linear trend, and $r^2$ captures the variation around the linear slope.
$\UH^{\mathrm{Tr}}$ is elevated on the incorrect trace across $\mu_{\mathrm{early}}$, $\mu_{\mathrm{mid}}$, and $\mu_{\mathrm{late}}$ and the incorrect slopes fail to decrease as cleanly as the correct ones, consistent with the analysis of \autoref{sec:analysis:features}.

The model correctly computes the relevant probabilities ($50\%$ for the first event, $25\%$ for two consecutive odd rolls) for both sequences and correctly considers the same two interpretations of \textit{``how much more likely''}: the \textit{difference} and the \textit{ratio}.
In the incorrect sequence, the model commits to using the ratio, shown in the highlighted sentence $31$ of \autoref{fig:gsm-symbolic} which corresponds to the highlighted token number $500$;
this results in the model incorrectly outputting $200\%$ instead of $25\%$ as the final answer.
The correct trace peaks at the multiplicative-probability composition, while the incorrect trace shows a local elevation when committing to the ratio interpretation, with the signal reaching roughly $2.4\times$ the average of the surrounding $200$-token window.
For the correct trace, there is no peak at the mention of the ratio, but at the point of committing to the difference.
Importantly, focusing on the first semantically observable error in the generated reasoning aligns only with a single peak of the trace, at token $500$.
Many of the earlier peaks could be relevant for error-detection.

Both trajectories additionally show their largest relative peaks at structural boundaries---the trace opening and the post-think \textit{``Final Answer''} section header---which we read as tokenization-driven gradient elevation rather than reasoning content; against this baseline the smaller ``reasoning''-content peaks are the ones that distinguish the two traces and that flag the specific ``reasoning'' steps at which the incorrect trace shows errors.


\section{Discussion}
\label{sec:application}

The UQ lens reveals a stark difference between generative processes resulting in correct and incorrect final answers, answering \textbf{RQ1} in the affirmative: uncertainty trace profiles predict correctness with AUROC up to $0.807$, outperforming \textit{Self-Certainty} \citep{kang-2025-scalablebestofn} and \textit{Circuit-Based Reasoning Verification} \citep{zhao-2025-verifyingchainofthought}, and one can predict correctness with AUROC up to $0.801$ using only the first $300$ tokens.

UQ provides an interpretative lens through which to understand the process of autoregressive generation that is currently being labeled as ``reasoning''.
Correct and incorrect traces are distinguishable across the studied models, and the analysis shows that the demarcation is made based on a relative difference in the slope and linearity of the entropy, $\UH$, and Bernoulli variance, $\UA$, with a steeper and more linear slope being indicative of a trace being incorrect.
Notably, the epistemic uncertainty presents a complementary behavior, being smaller for incorrect traces than for correct traces.
Together, the aleatoric trajectory and the epistemic dissociation answer \textbf{RQ2}: the discriminating signals lie in the joint behavior of the two channels, not in either taken in isolation.
Under the classical interpretation of aleatoric and epistemic uncertainty discussed in \autoref{sec:bg:uncertainty-in-ml}, the most intuitive failure mode couples the two measures: an out-of-distribution input should yield elevated epistemic uncertainty, and a difficult input should yield elevated aleatoric uncertainty.
Our findings deviate from this interpretation: incorrect traces show \textit{elevated} aleatoric and \textit{depressed} epistemic uncertainty.
The model is uncertain during generation, yet the wrong answer it produces is seemingly well-supported by training data;
aleatoric and epistemic thus carry separate diagnostic signals: neither alone reveals this failure mode.

\paragraph{Implications.}
\label{sec:discussion:implications}
First, the predictive power of uncertainty trace profiles offers a practical signal for early correctness estimation, in both LMs and RMs.
This has potential applications in selective generation and test-time scaling that aim to terminate or resample traces prior to completion.
Second, the observed dissociation between aleatoric and epistemic measures of uncertainty points to a characterization of failure modes that simpler measures miss.
Third, our case study illustrates failures that uncertainty-based analysis can identify but semantic analysis of the trace cannot, establishing the uncertainty lens as an analytical tool that yields complementary insights to semantic inspection.

\paragraph{Limitations and future work.}
\label{sec:discussion:limitations}

We propose a method for analyzing ``reasoning'' based on temporal dynamics of the generative process, which have for the most part been ignored by prior work.
Our method and analysis rely on interpretations of aleatoric and epistemic uncertainty estimates that are commonly used and well-founded, but nevertheless rest on approximations that are defensible in the overparameterized regime, yet not strictly verifiable.
Further, the long sequences generated by RMs sometimes exceed the maximum length of $2048$ tokens set for the experiments, potentially distorting results;
we mitigate this through the filtering procedure described in \autoref{sec:methods:procedure}.
The results differ in consistency between GSM8K, an older dataset which is likely to be in pre-training data, and ProntoQA, which is more recent and possibly more indicative of out-of-distribution ``reasoning'' due to the nature of the examples.
We emphasize that this paper presents a proof-of-concept for the informativeness of uncertainty measures across ``reasoning'' traces, which we show on five models ($0.5$B-$8$B parameters) spanning the LM to RM spectrum and two commonly used datasets for mathematical and logical reasoning.
Future work should expand the evaluation to other types of ``reasoning'' tasks, different model architectures and sizes, as well as various out-of-distribution shifts, to further generalize our results.
\section{Conclusion}
\label{sec:conclusion}

We have proposed treating LM ``reasoning'' traces as time series of uncertainty estimates, summarizing each trace by an uncertainty trace profile that captures the level and dynamics of uncertainty over the course of generation.
The uncertainty trace profile is strongly predictive of correctness across five models and two datasets, with AUROC up to $0.807$ on full traces and $0.801$ using only the first few hundred tokens, a significant improvement over prior approaches.
We further show that correct and incorrect traces differ systematically under this lens, and that the epistemic and aleatoric channels carry diagnostic signal in opposite directions, indicating that the two cannot be summarized by a single measure.
Taken together, these findings indicate that uncertainty quantification across ``reasoning'' traces captures substantial information about the dynamics of the process, and offers a lens for studying it that semantic inspection alone does not provide.

\section*{Acknowledgments}

This work was supported by the Danish Data Science Academy, which is funded by the Novo Nordisk Foundation (NNF21SA0069429) and VILLUM FONDEN (40516). We further acknowledge the support for BP through the ERC Consolidator Grant DIALECT 101043235.

\bibliographystyle{abbrvnat}
\bibliography{bibliography}
\newpage

\appendix
\section{Intuitions for the Uncertainty Types}
\label{app:uncertainty-intuitions}

The three uncertainty types defined in \autoref{sec:bg:uncertainty-in-ml} can be unfamiliar on first reading.
This appendix offers worked intuitions for each, grouped by the aleatoric--epistemic distinction (\autoref{app:intuitions:ae}) and the committal--distributional distinction within aleatoric uncertainty (\autoref{app:intuitions:cd}).

\subsection{Aleatoric and Epistemic Uncertainty}
\label{app:intuitions:ae}

Consider trying to predict the outcome of a coin flip for a coin you have just been handed.
You have flipped it a few times to estimate whether it is fair, but you remain unsure.
Two distinct sources contribute to your uncertainty about the next flip.

\paragraph{Aleatoric (irreducible) uncertainty.}
A coin flip is, in principle, a random event.
Even if you knew with perfect certainty that the coin was exactly fair, you still could not predict the outcome of any particular flip with confidence;
the randomness is intrinsic to the task.
This is aleatoric uncertainty.
Flipping the coin more times to learn about its bias does not reduce it.
In the language of \autoref{sec:bg:uncertainty-in-ml}, it is the noise in $y$ given $x$ that no parameterization $\theta$ could remove.

\paragraph{Epistemic (knowledge-based) uncertainty.}
You have flipped the coin only a handful of times, and you are not yet sure whether it is fair, slightly biased toward heads, or slightly biased toward tails.
This second source of uncertainty is not about the coin flip itself but about your belief regarding the coin's underlying bias.
Someone who had flipped the coin a thousand times would be far more certain about its bias and therefore better calibrated about the next flip, even though the flip itself remains as random as ever.
This difference is epistemic uncertainty: it is reducible in principle, by gathering more evidence.
In the language of \autoref{sec:bg:uncertainty-in-ml}, it is the variability in the predictive distribution across plausible parameterizations $\theta$ given the observed data.

\subsection{Committal and Distributional Aleatoric Uncertainty}
\label{app:intuitions:cd}

Within the aleatoric component, our work distinguishes two sub-types that capture different aspects of how a model commits to a prediction.
Consider a language model deciding which token to generate next, with a vocabulary of $50{,}000$ tokens.
Three illustrative cases bring out the distinction:

\paragraph{Case 1: One token clearly dominates.}
The model assigns probability $0.95$ to the token \texttt{the}, with the remaining $0.05$ spread roughly evenly across the other $49{,}999$ tokens.
\textit{Committal} uncertainty is low: the Bernoulli variance $0.95 \cdot 0.05 = 0.0475$ is close to zero, indicating that the model is firmly committed to \texttt{the}.
\textit{Distributional} uncertainty is also low: the entropy is small because almost all the mass sits on a single token.
This is the regime we call \textbf{concentration}: the top token wins by being likely.

\paragraph{Case 2: One token nominally wins, but mass is broadly distributed.}
The model assigns probability $0.15$ to \texttt{the}, with the remaining $0.85$ spread roughly evenly across $1{,}000$ other tokens that each receive roughly $0.00085$.
\textit{Committal} uncertainty is high: the Bernoulli variance $0.15 \cdot 0.85 = 0.13$ is substantial relative to its maximum of $0.25$, indicating that the model's top choice carries little of the total probability mass.
\textit{Distributional} uncertainty is also high: the entropy is large because mass is spread across many alternatives.
This is the regime we call \textbf{elimination}: the token \texttt{the} wins not by being likely in absolute terms, but by being marginally more probable than each of many alternatives that are individually even less probable.

\paragraph{Case 3: Two tokens roughly tied.}
The model assigns probability $0.5$ to \texttt{the} and $0.5$ to \texttt{a}, with negligible mass elsewhere.
\textit{Committal} uncertainty is at its maximum: the Bernoulli variance $0.5 \cdot 0.5 = 0.25$ reflects that the top token wins by an arbitrarily small margin.
\textit{Distributional} uncertainty is low: the entropy is just $\log_2 2 = 1$ bit, because only two tokens carry mass.
This case sits at the boundary of the elimination regime, with the elimination occurring among only two alternatives rather than many; with so little of the mass distributed, distributional uncertainty stays low even though committal uncertainty is maximal.

The cases above show that the two measures can move independently: the elimination regime (Case 2) and a near-tied two-way contest (Case 3) both produce high committal uncertainty, but only Case 2 also produces high entropy.
The same low-entropy reading can therefore correspond to confident commitment (Case 1) or to a near-tied two-way contest (Case 3), and only committal uncertainty distinguishes them.

\section{Extracting the Final Answer, $\hat{y}$}
\label{app:splitting}


Computing answer uncertainty $\UTA$ requires identifying, for each generation, the span $\hat{y}$ corresponding to the model's final answer.
We split each model's output into a trace section and an answer section using model-specific rules, since different model families structure their outputs differently.
In all cases the split is a position in the generated string, with the concatenation of the two sections reproducing the original output exactly.
The strategies were developed iteratively by inspecting splits on samples of correct and incorrect generations from each model-dataset combination, with the additional check that trace and answer sections concatenate back to the original output character-for-character.

\paragraph{Reasoning models (DeepSeek R1, Qwen 3).}
Both models wrap their internal reasoning in explicit \texttt{<think>}\,\ldots\,\texttt{</think>} tags.
We first locate the closing \texttt{</think>} tag and then search the post-think content for the latest occurrence of an answer-declaration pattern, such as ``Final Answer'', ``\#\# Final Answer'', ``Answer:'', or ``Therefore, the answer is''.
The split is placed immediately after this declaration, so the answer section contains only the answer itself and not the surrounding markup.
By construction (see \autoref{sec:methods:procedure}), every retained reasoning model trace contains a closing \texttt{</think>} tag and a recognizable answer declaration: traces missing these were either regenerated with an additional $512$-token budget or filtered out if the extended budget still did not yield an answer.

\paragraph{Non-reasoning models (Qwen 2.5, Llama 3.1, Llama 3.2).}
These models lack explicit reasoning sections and instead structure their outputs as plain text in which the final answer is conventionally stated in a concluding paragraph.
We split at the last paragraph break (an empty line, matched as \texttt{\textbackslash n\textbackslash s*\textbackslash n}), taking everything after this break as the answer.
If no paragraph break is found, we fall back to the last single newline; if neither is present and cannot reliably split the section we exclude the sample from the $\UTA$ computation.

\section{Trace Features}
\label{app:trace-feats}

Based on the generated traces and the uncertainty measure we compute the following metrics for each type of uncertainty. Recall that the uncertainty types are the entropy of the predictive distribution, an approximation to the aleatoric uncertainty and finally an approximation to the epistemic uncertainty as described in \autoref{sec:methods:uncertainty}.

\begin{table}[h!]
\centering
\begin{tabularx}{\textwidth}{llX}
\hline
\textbf{Feature} & \textbf{Symbol} & \textbf{Description} \\
\hline
\multicolumn{3}{l}{\textit{Static Features}} \\
Early mean & $\mu_{\mathrm{early}}$ & Arithmetic mean of the first $25\%$ trace. \\
Middle mean & $\mu_{\mathrm{mid}}$ & Arithmetic mean of the middle $50\%$ trace. \\
Late Mean & $\mu_{\mathrm{late}}$ & Arithmetic mean of the last $25\%$ trace. \\
\hline
\\[-1.5ex]
\multicolumn{3}{l}{\textit{Dynamic Features}} \\
Slope & $\mathbf{m}$ & Slope of a linear trend fitted on a trace. \\
Linear fit & $r^2$ & Goodness-of-fit of the linear trend. \\
\hline
\end{tabularx}
\caption{Overview of features extracted from every uncertainty trace.}
\label{tab:trace-features}
\end{table}

\clearpage
\section{Experimental Setup}
\label{app:experimental-setup}

This appendix lists implementation details that are needed for reproducibility but kept out of the main text.
The model lineup, datasets, generation protocol, uncertainty measures, and feature definitions themselves are described in \autoref{sec:methods:procedure} and the trace/answer splitting in the section above.

\subsection{Sample Retention}
\label{app:retention}

Two filters are applied to each generated sample at load time, before any analysis:
\begin{enumerate}
    \item \textbf{\texttt{contains\_answer} check.} A regex-based extractor attempts to pull a final answer from the answer portion of each generation. Generations for which extraction fails are dropped.
    \item \textbf{LLM correctness re-evaluation.} For samples whose extracted answer differs from the reference under the regex matcher, we query an LLM auditor (\texttt{gpt-5-2025-08-07}) to re-evaluate correctness under a more lenient semantic match (e.g., $0.5$ vs $1/2$, mathematically equivalent expressions). When the auditor confirms correctness, we flip \texttt{is\_correct} to \texttt{True}; otherwise we keep the original label.
\end{enumerate}
For Qwen 3 and DeepSeek R1, traces that exhaust the $2048$-token budget without emitting a closing \texttt{</think>} tag are regenerated with a forced \texttt{</think>} injection and an additional $512$-token continuation; traces still missing a final answer after the regeneration are dropped.

\subsection{Predictive Modeling}
\label{app:predictive}

The two classifiers in \autoref{sec:analysis:correctness} are instantiated as
\begin{itemize}
    \item \texttt{LogisticRegression(max\_iter=1000, random\_state=42)} with default L2 regularization ($C=1$);
    \item \texttt{GradientBoostingClassifier(n\_estimators=100, max\_depth=3, random\_state=42)},
\end{itemize}
both with \texttt{StandardScaler} fit on the training fold inside the CV loop. Cross-validation uses \texttt{StratifiedKFold(n\_splits=5, shuffle=True, random\_state=42)}, constructed independently per (model, dataset) combination.

\subsection{Compute and Reproducibility}
\label{app:compute}

All classifiers and CV splits are seeded with \texttt{random\_state=42}.
Trace generation is greedy decoding, deterministic given the model weights and tokenizer.
All experiments are run on H100 GPUs.
Source code is released alongside the paper.

\clearpage
\section{Additional Analysis}
\label{app:analysis}

\subsection{AUROC by Uncertainty Type}
\label{app:auroc-by-type}

We distinguish between type-channel combinations, and observe that the trace-level uncertainties, $\UT$, are generally more predictive than answer-level uncertainties, $\UTA$.
Additionally, both aleatoric and epistemic uncertainty carry relevant predictive power as seen in columns $\UH$ and $\UE$.


\begin{table}[h!]
\centering
\small
\begin{tabularx}{\textwidth}{l l *{4}{>{\centering\arraybackslash}X}}
\toprule
& & \multicolumn{2}{c}{GSM8K} & \multicolumn{2}{c}{ProntoQA} \\
\cmidrule(lr){3-4} \cmidrule(lr){5-6}
& & LR & GB & LR & GB \\
\midrule
\multirow{5}{*}{\rotatebox[origin=c]{90}{$\UT$}}
& Llama 3.1   & 0.781 {\tiny ±0.034} & 0.741 {\tiny ±0.043} & 0.777 {\tiny ±0.057} & 0.712 {\tiny ±0.118} \\
& Llama 3.2   & 0.754 {\tiny ±0.011} & 0.747 {\tiny ±0.033} & 0.538 {\tiny ±0.035} & 0.553 {\tiny ±0.046} \\
& Qwen 2.5    & 0.787 {\tiny ±0.027} & 0.777 {\tiny ±0.023} & 0.521 {\tiny ±0.036} & 0.501 {\tiny ±0.013} \\
& DeepSeek R1 & 0.765 {\tiny ±0.032} & 0.753 {\tiny ±0.043} & 0.686 {\tiny ±0.051} & 0.647 {\tiny ±0.076} \\
& Qwen 3      & 0.722 {\tiny ±0.045} & 0.692 {\tiny ±0.029} & 0.674 {\tiny ±0.073} & 0.578 {\tiny ±0.044} \\
\midrule
\multirow{5}{*}{\rotatebox[origin=c]{90}{$\UTA$}}
& Llama 3.1   & 0.682 {\tiny ±0.030} & 0.670 {\tiny ±0.019} & 0.669 {\tiny ±0.124} & 0.714 {\tiny ±0.111} \\
& Llama 3.2   & 0.675 {\tiny ±0.017} & 0.660 {\tiny ±0.027} & 0.544 {\tiny ±0.050} & 0.529 {\tiny ±0.047} \\
& Qwen 2.5    & 0.755 {\tiny ±0.030} & 0.716 {\tiny ±0.033} & 0.566 {\tiny ±0.057} & 0.481 {\tiny ±0.048} \\
& DeepSeek R1 & 0.642 {\tiny ±0.055} & 0.629 {\tiny ±0.048} & 0.660 {\tiny ±0.088} & 0.638 {\tiny ±0.069} \\
& Qwen 3      & 0.685 {\tiny ±0.053} & 0.616 {\tiny ±0.043} & 0.619 {\tiny ±0.103} & 0.502 {\tiny ±0.113} \\
\bottomrule
\end{tabularx}
\caption{AUROC scores (5-fold CV, curated features) across models and classifiers for GSM8K and ProntoQA. Values shown as mean ± std.
LR = Logistic Regression, GB = Gradient Boosting.}
\label{tab:auroc-split}
\end{table}

\begin{table}[h!]
\centering
\small
\begin{tabularx}{\textwidth}{l l *{6}{>{\centering\arraybackslash}X}}
\toprule
& & \multicolumn{3}{c}{GSM8K} & \multicolumn{3}{c}{ProntoQA} \\
\cmidrule(lr){3-5} \cmidrule(lr){6-8}
& & $\UH$ & $\UA$ & $\UE$ & $\UH$ & $\UA$ & $\UE$ \\
\midrule
\multirow{5}{*}{\rotatebox[origin=c]{90}{$\UTA$}}
& Llama 3.1   & 0.682 {\tiny ±0.033} & 0.630 {\tiny ±0.030} & 0.572 {\tiny ±0.035} & 0.629 {\tiny ±0.121} & 0.603 {\tiny ±0.150} & 0.523 {\tiny ±0.103} \\
& Llama 3.2   & 0.634 {\tiny ±0.031} & 0.582 {\tiny ±0.033} & 0.587 {\tiny ±0.033} & 0.537 {\tiny ±0.042} & 0.516 {\tiny ±0.063} & 0.529 {\tiny ±0.063} \\
& Qwen 2.5    & 0.718 {\tiny ±0.034} & 0.655 {\tiny ±0.009} & 0.678 {\tiny ±0.026} & 0.571 {\tiny ±0.045} & 0.539 {\tiny ±0.057} & 0.496 {\tiny ±0.025} \\
& DeepSeek R1 & 0.613 {\tiny ±0.048} & 0.549 {\tiny ±0.052} & 0.568 {\tiny ±0.031} & 0.693 {\tiny ±0.103} & 0.477 {\tiny ±0.102} & 0.658 {\tiny ±0.066} \\
& Qwen 3      & 0.657 {\tiny ±0.020} & 0.659 {\tiny ±0.063} & 0.636 {\tiny ±0.045} & 0.525 {\tiny ±0.049} & 0.679 {\tiny ±0.133} & 0.555 {\tiny ±0.091} \\
\midrule
\multirow{5}{*}{\rotatebox[origin=c]{90}{$\UT$}}
& Llama 3.1   & 0.717 {\tiny ±0.062} & 0.705 {\tiny ±0.060} & 0.778 {\tiny ±0.043} & 0.811 {\tiny ±0.052} & 0.776 {\tiny ±0.047} & 0.712 {\tiny ±0.075} \\
& Llama 3.2   & 0.684 {\tiny ±0.020} & 0.679 {\tiny ±0.020} & 0.757 {\tiny ±0.027} & 0.532 {\tiny ±0.051} & 0.520 {\tiny ±0.057} & 0.512 {\tiny ±0.041} \\
& Qwen 2.5    & 0.788 {\tiny ±0.031} & 0.786 {\tiny ±0.030} & 0.738 {\tiny ±0.020} & 0.544 {\tiny ±0.046} & 0.562 {\tiny ±0.050} & 0.517 {\tiny ±0.047} \\
& DeepSeek R1 & 0.703 {\tiny ±0.026} & 0.615 {\tiny ±0.024} & 0.736 {\tiny ±0.036} & 0.664 {\tiny ±0.060} & 0.679 {\tiny ±0.058} & 0.504 {\tiny ±0.064} \\
& Qwen 3      & 0.723 {\tiny ±0.049} & 0.717 {\tiny ±0.043} & 0.549 {\tiny ±0.084} & 0.600 {\tiny ±0.052} & 0.558 {\tiny ±0.049} & 0.528 {\tiny ±0.087} \\
\bottomrule
\end{tabularx}
\caption{AUROC scores (Logistic regression with 5-fold CV) by uncertainty type across models for GSM8K and ProntoQA. Values shown as mean ± std.}
\label{tab:auroc-uncertainty-types}
\end{table}

\clearpage

\end{document}